\documentclass[12pt]{article}
\usepackage[margin=1in]{geometry}
\usepackage{amsmath, amssymb}
\usepackage{booktabs}
\usepackage{caption}
\usepackage{graphicx}

\title{Neural Networks with Local Converging Inputs for Efficient Options Pricing Models}
\author{Harris Cobb\footnotemark[1], Wenbo Hao\footnotemark[1], and Yingjie Liu\footnotemark[1]}
\date{}

\begin{document}

\maketitle
\footnotetext[1]{
School of Mathematics, Georgia Institute of Technology,
Atlanta, GA 30332 ({\tt hcobb7@gatech.edu, whao36@gatech.edu, yingjie@math.gatech.edu}).}

\begin{abstract}
We present a novel application of Neural Networks with Local Converging Inputs (NNLCI) to improve the efficiency of existing numerical methods for pricing multi-asset options. The most concise input format for NNLCI has been introduced, offering substantial convenience and efficiency. NNLCI uses a neural network to locally correct solutions from a coarse mesh and a refined mesh (relative to the coarse one), requiring only a minimal amount of high-fidelity training data. We demonstrate this approach on cash-or-nothing options under the Black–Scholes equation in one, two, and three spatial dimensions, and on single-asset down-and-out barrier call options under the Heston stochastic-volatility model (whose pricing PDE is two-dimensional in the spot price \(S\) and the instantaneous variance \(v\)). In each case, NNLCI reduces the root-mean-square error (RMSE) of the refined-mesh numerical solution by a factor of approximately 4–12 on test sets, even when the neural network is trained on only a small subset of parameter combinations. These results demonstrate that NNLCI significantly reduces computational requirements for high-dimensional problems in real-time options trading and risk management, offering low training costs and strong generalization ability.
\end{abstract}

\section{Introduction and Motivation}

Numerical methods for partial differential equations (PDEs) play a central role in option pricing. An \emph{option} is a financial derivative that grants its holder the right, but not the obligation, to buy or sell an asset at a predetermined \emph{strike price} \(K\) at or before a specified \emph{maturity time} \(T\). Fast and accurate option pricing is crucial in practice: traders require near-instantaneous pricing in response to changing market conditions, especially for multi-asset options where numerical complexity grows exponentially in the number of underlying assets. Traditional finite-difference or finite-element solvers can be computationally expensive for high-dimensional PDEs, limiting their practical use in real-time trading and risk management.

We apply \emph{Neural Networks with Local Converging Inputs} (NNLCI) to accelerate numerical PDE solvers for option pricing. NNLCI was originally developed to accurately predict solutions involving shocks, contacts and their interactions for hyperbolic PDEs by locally combining coarse- and refined-mesh data within a neural network~\cite{Huang2023NNLCI1D_CCP,Huang2023NNLCI2D_CCP}. Since then, it has been extended to electromagnetic waves scattered by complex perfect electric conductors \cite{Cobb2023Maxwell}, to unstructured grids \cite{Ding2024NNLCI}, and to elliptic systems defined on multiple subdomains~\cite{Lee2024PNPic}. Although NNLCI has demonstrated the potential to significantly increase accuracy on many PDEs, it does not perform equally well for all problems; for example, it is not effective for systems exhibiting chaotic behavior~\cite{Cobb2025NNLCI}. Here, we adapt NNLCI to parabolic PDEs arising in finance: the Black–Scholes equation for European-style \emph{cash-or-nothing call options} and the Heston stochastic-volatility model for \emph{down-and-out barrier call options}. A cash-or-nothing call option pays a fixed cash amount if the underlying asset price \(S_T\) at maturity exceeds the strike \(K\), and pays zero otherwise. A down-and-out barrier call option pays \(\max\{S_T - K,0\}\) if the underlying asset price has remained above a barrier \(B\) throughout \([0,T]\); if the asset ever touches or crosses \(B\) before maturity, the option expires worthless.

The motivation for using NNLCI in option pricing is twofold. First, multi-asset options (with 2 or more correlated underlyings) lead to high-dimensional PDEs, making high-resolution solvers extremely expensive. NNLCI uses only a small set of high-fidelity (reference-mesh) solutions as training data; once trained, the network can quickly map a coarse-mesh solution together with a solution on a mesh refined relative to that coarse grid to an approximation whose accuracy is substantially higher than that of the coarse solution, at far lower cost than computing a reference-mesh solution at inference time. Second, NNLCI exhibits strong generalization: a trained network can transfer across different PDE parameters, geometries, and initial and boundary conditions~\cite{Cobb2025NNLCI}, because it focuses on local dynamics rather than modeling the entire PDE solution operator end-to-end. As a result, one can achieve dramatic accuracy improvements with minimal training data and a simple feed-forward network architecture.

We organize this paper as follows. Section ~\ref{sec:related} outlines related work. Section ~\ref{sec:method} reviews the NNLCI methodology. Section~\ref{sec:experiments} presents numerical experiments on cash-or-nothing options (1D, 2D, 3D) and on down-and-out barrier calls under the Heston model. Section~\ref{sec:conclusion} concludes and discusses future directions.

\section{Related Work}
\label{sec:related}
\subsection{Classical PDE and Monte Carlo Methods}

Option prices can be obtained by solving an associated pricing PDE. Numerical methods for solving PDEs are already well-developed and span a broad spectrum, including finite‐difference (e.g.\ Crank–Nicolson), finite‐element, finite‐volume, spectral methods, and optimization-based iterative solvers ~\cite{Wilmott1998,Ikonen2008,LeVeque1992,Boyd2001,Meng2025}. Accuracy and stability for interface and advection-dominated problems have been improved using level set methods and back-and-forth error compensation and correction (BFECC), applied in contexts such as Maxwell's equations, semi-Lagrangian schemes, and multiphase flows~\cite{Dupont2003LevelSetBFECC,Dupont2007BFECC,FlowFixer2005,Wang2019BFECCMaxwell,Guo2022Maxwell}. For high-dimensional problems, sparse grids~\cite{Zhu2002} and domain decomposition~\cite{Caflisch2000} offer partial relief from the curse of dimensionality on PDE grids. One can also bypass the PDE and estimate the risk-neutral expectation directly by Monte Carlo path sampling---including quasi-MC and multilevel MC---whose cost grows far more mildly with dimension and is especially attractive for multi-asset and path-dependent payoffs~\cite{Glasserman2003,Giles2008}.

\subsection{Neural PDE Solvers}

Feed‐forward networks have learned option prices from simulated data~\cite{Hutchinson1994}. Neural networks that enforce PDE residuals (e.g., Dissanayake and Phan-Thien~\cite{DiPh94}, physics‐informed neural networks (PINNs)~\cite{Raissi2019}) often fail to capture steep payoff features without dense sampling and typically incur high computational complexity. Operator‐learning models—Chen and Chen~\cite{ChenChen95}, DeepONet~\cite{Lu2019DeepONet} and Fourier Neural Operator (FNO)~\cite{Li2020FNO}—approximate the parameter‐to‐solution map, though they typically require extensive high‐fidelity data~\cite{Jin2021,Fournier2022}. Gradient information has since been incorporated to alleviate this data burden~\cite{Qiu2024,OLearyRoseberry2024}. Variants such as Physics-Informed Neural Operators (PINO)~\cite{Li2024PINO} further embed physical constraints to enhance generalization and computational efficiency. 

\subsection{NNLCI Developments}
Neural Networks with Local Converging Inputs were introduced for one‐dimensional hyperbolic conservation laws to predict reference‐quality solutions using two nested approximations on small stencils \cite{Huang2023NNLCI1D_CCP}. Extensions to two‐dimensional Euler flows \cite{Huang2023NNLCI2D_CCP}, electromagnetic scattering \cite{Cobb2023Maxwell}, and unstructured‐mesh supersonic flows \cite{Ding2024NNLCI} demonstrated NNLCI’s versatility. Lee et al.\ adapted NNLCI to a 1D Poisson–Nernst–Planck ion‐channel system with interfaces, achieving substantial speedups over fine‐mesh FEM by learning interfacial jumps implicitly \cite{Lee2024PNPic}. Cobb further extended NNLCI to quantum many-body problems~\cite{Cobb2025NNLCI}. NNLCI’s highly localized input reduces both network size and training data requirements, enabling stronger generalization.

\subsection{Neural Network Methods for Financial PDEs}
Although PINNs and operator nets have been applied to Black–Scholes and Heston equations \cite{Jin2021, Sirignano2018}, no prior work employs NNLCI’s nested‐grid, local‐patch paradigm for option pricing. Our approach fills this gap by integrating coarse‐ and refined-mesh finite‐difference/finite‐element approximations into a local neural correction, achieving high accuracy with minimal high‐fidelity samples in multi‐asset and barrier settings.

\section{Neural Networks with Local Converging Inputs}
\label{sec:method}

NNLCI is designed to enhance traditional numerical solvers for PDEs by learning a local correction from coarse- and refined-mesh solutions toward a reference (highest-resolution) mesh solution. The key ideas are:
\begin{itemize}
    \item Construct three meshes of different resolutions: a \emph{coarse mesh} used to crop the first neural‑network input, a \emph{refined mesh} (obtained by doubling the resolution of the coarse mesh and used to crop the second input), and a \emph{reference mesh} (the highest-resolution mesh) serving as the ground‑truth target for the network’s output.
    \item At each spatio-temporal location of interest (a \emph{collocation point}), gather the computed PDE solutions on both coarse and refined meshes at grid points captured by a local patch centered at the collocation point. 
    \item Feed this pair of local-patch values to a neural network, which outputs a corrected solution at the center of the patch that aims to match the reference-mesh solution.
\end{itemize}

Importantly, the neural network \emph{does not} receive as inputs (i) the spatio-temporal coordinates, (ii) PDE hyperparameters (e.g., volatility \(\sigma\), interest rate \(r\), mean-reversion \(\kappa\), etc.), or (iii) boundary/initial-condition metadata because the information of these parameters is contained in the local, converging inputs of PDE solutions. By contrast, many existing neural PDE solvers treat these parameters explicitly, which leads to more complex architectures, greater training data demands, and reduced generalization capability. NNLCI instead treats these quantities \emph{implicitly} by focusing on the local dynamics of the solutions.

Formally, let \(\mathcal{P}\) denote a local patch centered at \((t_i, x_i)\) on the coarse mesh and the same patch on the refined mesh. Let
\[
u^C_{\,\mathcal{P}} \;=\; \{u^C(t_j,x_k)\,\bigm|\,(t_j,x_k)\in \mathcal{P}\}\,, 
\qquad
u^F_{\,\mathcal{P}} \;=\; \{u^F(t_j,x_k)\,\bigm|\,(t_j,x_k)\in \mathcal{P}\}
\]
denote the sets of computed solutions on coarse (C) and refined (F) meshes, respectively, restricted to \(\mathcal{P}\). The neural network
\[
\mathcal{N}\bigl(u^C_{\,\mathcal{P}},\,u^F_{\,\mathcal{P}}\bigr)
\;\approx\;
u^R(t_i,x_i)\,,
\]
is trained to predict the reference-mesh value \(u^R(t_i,x_i)\). The number of input nodes to \(\mathcal{N}\) is determined by the local-patch size, whose choice we discuss after the following subsection.

\subsection{Handling Parameter Implicitness}

Because NNLCI omits direct encoding of spatiotemporal coordinates, PDE coefficients, and boundary/initial data, there is no theoretical guarantee that a local coarse solution patch uniquely determines the reference solution; distinct global solutions may share identical local coarse solution patterns (a manifestation of non-injectivity), though using local, converging patches of coarse solutions substantially reduces this risk.  This issue is acute in chaotic systems, where small differences in global state lead to large deviations locally. Indeed, experiments in ~\cite{Cobb2025NNLCI} show that NNLCI may not be effective for the double pendulum and three-body problems: NNLCI failed to converge beyond a short time horizon owing to sensitive dependence on initial conditions. We therefore restrict our focus to \emph{non-chaotic} PDEs (e.g., parabolic equations for diffusion, convection–diffusion, and option pricing), where local solution patterns more reliably encode the relevant global information.

In exchange for forgoing global uniqueness, NNLCI benefits from:
\begin{itemize}
    \item \textbf{Data efficiency.} Only a small amount of high-fidelity training data is required, which is important because data generation is often one of the most time-consuming stages for other machine-learning PDE models such as neural operators.
    \item \textbf{Architectural simplicity.} Because the network learns a local correction map rather than a global solution operator, the target map is comparatively easy to learn; the low-dimensional local inputs also keep the number of trainable parameters small, making the network highly memory-friendly.
    \item \textbf{Reuse without retraining.} Once trained, the same network can be applied under moderate changes in PDE parameters, spatial locations, boundary conditions, and geometry without retraining; retraining is needed primarily when the underlying PDE itself changes.
\end{itemize}

\subsection{Choice of Local Patch Size}

The local-patch size is an important hyperparameter: including more points tends to improve accuracy, but also increases the complexity of the learned map and thereby the training-data requirement and training difficulty. The size that best balances cost and accuracy depends on the PDE type and mesh structure.

For hyperbolic equations (e.g., Euler and Maxwell's equations), patches are typically chosen from the local domain of dependence (e.g., in 1D: \(\{(t_i,x_i),(t_{i-1},x_{i-1}), (t_{i-1},x_i), (t_{i-1},x_{i+1})\}\) on both coarse and refined meshes, for a total of 8 inputs). In two spatial dimensions, a \(3\times 3\) spatial patch (18 inputs total) sufficed for time-reversible Maxwell’s equation~\cite{Cobb2023Maxwell}. On unstructured meshes, a local mesh-size parameter \(\,h\) was appended as an additional input~\cite{Ding2024NNLCI}.

For the parabolic option-pricing PDEs considered here, a singleton patch consisting of only the collocation point---on each of the coarse and refined meshes---already yields a substantial improvement over the refined-mesh solution, because solutions are generally monotonic in each spatial variable (i.e., asset price). In particular:
\begin{itemize}
    \item \textbf{Structured uniform grid:} For cash-or-nothing options under Black–Scholes, using only the values \(\{u^C(t_i,x_i),\,u^F(t_i,x_i)\}\) at the same collocation point \((t_i,x_i)\) (2 inputs) provides substantial improvement.
    \item \textbf{Non-uniform/unstructured grid:} For the Heston model on a non-uniform mesh, one adds a local grid-size metric of the coarse mesh
    \[
      h_{\text{local}} \;=\; \sqrt{(h_{\text{left}} + h_{\text{right}})\,(h_{\text{up}} + h_{\text{down}})} 
    \]
    as the third input, so that \(\{u^C(t_i,x_i),\,u^F(t_i,x_i),\,h_{\text{local}}\}\) are fed to the network.  Here $ h_{\text{left}}, \;h_{\text{right}},\; h_{\text{up}} \;{\rm and} \; h_{\text{down}}$ denote the adjacent mesh spacings in the four coordinate directions.
\end{itemize}
In future work, we plan to test larger local patches to explore how the accuracy improves with the patch size; see Section~\ref{sec:conclusion} for further discussion.

\section{Numerical Experiments}
\label{sec:experiments}

We test NNLCI on two classes of options:
\begin{enumerate}
    \item \emph{Multi-asset cash-or-nothing call options} under the Black–Scholes PDE (spatial dimension \(N = 1,2,3\)).
    \item \emph{Down-and-out barrier call options} under the Heston stochastic-volatility model (two spatial dimensions: asset price \(S\) and variance \(v\)).
\end{enumerate}

\subsection{Cash-Or-Nothing Call Options under Black–Scholes}

\subsubsection{Model Formulation}

Under the \(N\)-asset Black–Scholes framework, let \(\mathbf{S} = (S^{(1)}, \dots, S^{(N)})\) be the vector of underlying asset prices. The PDE for a European-style cash-or-nothing call option that pays a fixed cash amount \(C\) if \(\min_{1\leq j\leq N} S^{(j)}_T > K\) and zero otherwise satisfies
\[
\frac{\partial u}{\partial t} \;+\; \sum_{j=1}^N r\,S^{(j)} \frac{\partial u}{\partial S^{(j)}} \;+\; \frac{1}{2} \sum_{j=1}^N \sigma_j^2 (S^{(j)})^2 \frac{\partial^2 u}{\partial (S^{(j)})^2} \;+\; \sum_{1\leq i<j \leq N} \rho_{ij}\,\sigma_i\,\sigma_j\,S^{(i)} S^{(j)} \frac{\partial^2 u}{\partial S^{(i)}\partial S^{(j)}} \;-\; r\,u \;=\; 0,
\]
for \((t,\mathbf{S})\in [0,T)\times [0,S_{\max}]^N\), with terminal condition
\[
u(T,\mathbf{S}) \;=\; 
\begin{cases}
C, & \text{if }\min_{1\le j\le N} S^{(j)} > K,\\
0, & \text{otherwise}.
\end{cases}
\]
Here:
\begin{itemize}
    \item \(r\) is the risk-free interest rate.
    \item \(\sigma_j\) is the volatility of \(S^{(j)}\).
    \item \(\rho_{ij}\) is the correlation between the Brownian motions driving \(S^{(i)}\) and \(S^{(j)}\).
    \item \(K = 100\) is the strike price and \(C = 100\) is the cash payout (both fixed throughout our experiments).
    \item We truncate the spatial domain at \(S_{\max} = 300\), so that each \(S^{(j)}\in[0,S_{\max}]\).
    \item On the faces \(S^{(j)}=0\) and \(S^{(j)}=S_{\max}\), we impose Dirichlet boundary conditions consistent with the cash-or-nothing payoff (in particular, \(u=0\) whenever any coordinate vanishes, and the far-field value follows from the closed-form asymptotics).
\end{itemize}

The $N$-dimensional Black–Scholes PDE for a cash-or-nothing call admits a closed-form solution for any positive integer 
$N$. Rather than computing a numerical reference solution on a fine grid, we directly use this exact formula to serve as our benchmark. This approach not only saves computational effort but also yields a precise measure of relative accuracy when comparing to more complex option-pricing models. 

\subsubsection{Numerical Setup}

We discretize \([0,T]\times [0,S_{\max}]^N\) with an alternating-direction implicit (ADI) finite-difference scheme of first-order accuracy in both time and space. The nested meshes below are used for all spatial dimensions \(N=1,2,3\); parameter sampling and network details for \(N=1,2\) follow, while the \(N=3\) sampling protocol is given later.
\begin{itemize}
    \item \textbf{Time to maturity:} \(T=1\).
    \item \textbf{Coarse mesh:} \(21\) uniform nodes in \emph{each} spatial direction and \(21\) time steps (so the spatial interior has \(19\) nodes per direction; for \(N=3\) the spatial grid is \(21\times 21\times 21\)).
    \item \textbf{Refined mesh:} \(41\) uniform nodes in \emph{each} spatial direction and \(41\) time steps (for \(N=3\), \(41\times 41\times 41\)), nested so that every coarse node coincides with a refined node.
    \item \textbf{Collocation points:} For each parameter tuple, collocation points are the coarse time--space nodes with interior spatial location (\(19^N\) spatial sites). At each collocation point \((t_i,\mathbf{S}_i)\) the two network inputs are the coincident-node values \(\{u^C(t_i,\mathbf{S}_i),\,u^F(t_i,\mathbf{S}_i)\}\); no interpolation from the refined mesh is used.
    \item \textbf{Neural network (\(N=1,2\)):} Two hidden layers of 15 neurons (\(N=1\)) or 20 neurons (\(N=2\)), ReLU, Adam with learning rate \(1\times10^{-3}\); 1500 epochs for \(N=1\) and 2000 for \(N=2\).
    \item \textbf{Parameter grid (\(N=1\)):} Discretize \(\sigma_1\in[0.10,\,0.50]\) and \(r\in[0,\,0.05]\) into \(2^4=16\) equal subintervals each, yielding \((2^4+1)^2=17^2=289\) pairs \((\sigma_1,r)\).
    \item \textbf{Parameter grid (\(N=2\)):} Discretize \(\rho_{12}\in[-0.99,\,0.93]\), \(r\in[0,\,0.05]\), \(\sigma_1\in[0.10,\,0.50]\), and \(\sigma_2\in[0.10,\,0.50]\) into \(2^3=8\) equal subintervals each, yielding \((2^3+1)^4=9^4=6561\) tuples \((\rho_{12},r,\sigma_1,\sigma_2)\).
\end{itemize}

For \(N=1,2\), after the full parameter grid is fixed, we form training and testing sets by a \emph{training gap} \(g\). Concretely, if a parameter axis has been discretized into the ordered nodes \(\theta_0,\theta_1,\dots,\theta_M\), a gap of \(g\) keeps only the subsampled nodes \(\theta_0,\theta_g,\theta_{2g},\dots,\theta_M\) on that axis; the training set is then the Cartesian product of these subsampled axes, and all remaining full-grid tuples are held out for testing. Thus larger \(g\) means a sparser training set and a stricter test of generalization. 
For example, in 1D each of \(\sigma_1\) and \(r\) has \(M+1=17\) nodes; with \(g=2\) one keeps \(9\) nodes per axis, hence \(9^2=81\) training pairs and \(289-81=208\) testing pairs. The same rule yields \(5^2=25\), \(3^2=9\), and \(2^2=4\) training pairs for \(g=4,8,16\), respectively.
 We use $g\in \{2,\,4,\,8,\,16\}$ for $N=1$; and $g\in \{2,\,4,\,8\}$ for $N=2$. 

Throughout the cash-or-nothing experiments, errors are measured against the closed-form solution. For a set of parameter tuples (train or test), we report
\begin{itemize}
    \item \(\mathrm{RMSE}_{\text{refined}}\): root-mean-square error between the refined-mesh solution and the exact solution, aggregated over all collocation points and all parameter tuples in that set;
    \item \(\mathrm{RMSE}_{\text{NNLCI}}\): the same RMSE for the NNLCI output.
\end{itemize}
Subscripts ``train'' and ``test'' indicate which parameter set is used.

\subsubsection{1D Cash-Or-Nothing Experiments}

\paragraph{Results}

For each training gap \(g\in\{2,4,8,16\}\), we report \(\mathrm{RMSE}_{\text{refined}}\) and \(\mathrm{RMSE}_{\text{NNLCI}}\) on the training and testing parameter sets.

\begin{table}[h]
    \centering
    \caption{1D Cash-Or-Nothing: Training RMSEs}
    \label{tab:1d_train}
    \begin{tabular}{ccc}
        \toprule
        \textbf{Training Gap} & \textbf{Refined RMSE (Train)} & \textbf{NNLCI RMSE (Train)} \\
        \midrule
        2  & $2.119305 \times 10^{0}$ & $4.563925 \times 10^{-1}$ \\
        4  & $2.186143 \times 10^{0}$ & $4.599067 \times 10^{-1}$ \\
        8  & $2.433833 \times 10^{0}$ & $5.400022 \times 10^{-1}$ \\
        16 & $2.907363 \times 10^{0}$ & $5.908855 \times 10^{-1}$ \\
        \bottomrule
    \end{tabular}
\end{table}

\begin{table}[h]
    \centering
    \caption{1D Cash-Or-Nothing: Testing RMSEs}
    \label{tab:1d_test}
    \begin{tabular}{ccc}
        \toprule
        \textbf{Training Gap} & \textbf{Refined RMSE (Test)} & \textbf{NNLCI RMSE (Test)} \\
        \midrule
        2  & $1.999632 \times 10^{0}$ & $7.994198 \times 10^{-1}$ \\
        4  & $2.019259 \times 10^{0}$ & $6.231656 \times 10^{-1}$ \\
        8  & $2.022012 \times 10^{0}$ & $6.685388 \times 10^{-1}$ \\
        16 & $2.024100 \times 10^{0}$ & $6.801901 \times 10^{-1}$ \\
        \bottomrule
    \end{tabular}
\end{table}

\bigskip
\noindent
\emph{Discussion:} On the test sets, NNLCI reduces RMSE by a factor of roughly \(2.5\)--\(3.2\) relative to the refined-mesh solver (e.g., \(2.02/0.62\approx 3.2\) when \(g=4\)). The relative improvement remains stable as the training gap increases, indicating that only a small fraction of the parameter grid is needed for robust gains.

\subsubsection{2D Cash-Or-Nothing Experiments}

\paragraph{Results}

We vary the training gap \(g\in\{2,\,4,\,8\}\) and report the same refined-mesh and NNLCI RMSEs on the training and testing parameter sets.

\begin{table}[h]
    \centering
    \caption{2D Cash-Or-Nothing: Training RMSEs}
    \label{tab:2d_train}
    \begin{tabular}{ccc}
        \toprule
        \textbf{Training Gap} & \textbf{Refined RMSE (Train)} & \textbf{NNLCI RMSE (Train)} \\
        \midrule
        2 & $9.551897 \times 10^{0}$ & $2.059705 \times 10^{0}$ \\
        4 & $1.013214 \times 10^{1}$ & $2.963919 \times 10^{0}$ \\
        8 & $1.100761 \times 10^{1}$ & $3.051525 \times 10^{0}$ \\
        \bottomrule
    \end{tabular}
\end{table}

\begin{table}[h]
    \centering
    \caption{2D Cash-Or-Nothing: Testing RMSEs}
    \label{tab:2d_test}
    \begin{tabular}{ccc}
        \toprule
        \textbf{Training Gap} & \textbf{Refined RMSE (Test)} & \textbf{NNLCI RMSE (Test)} \\
        \midrule
        2 & $9.206242 \times 10^{0}$ & $1.937728 \times 10^{0}$ \\
        4 & $9.228365 \times 10^{0}$ & $2.516097 \times 10^{0}$ \\
        8 & $9.236010 \times 10^{0}$ & $2.311679 \times 10^{0}$ \\
        \bottomrule
    \end{tabular}
\end{table}

\bigskip
\noindent
\emph{Discussion:} On the test sets, NNLCI reduces RMSE by a factor of about \(4\)--\(5\) (e.g., \(9.21/1.94\approx 4.8\) when \(g=2\)). The gain is nearly independent of the training gap; even for \(g=8\), only \(2^4=16\) of the \(6561\) parameter tuples (\(\approx 0.24\%\)) are used for training.

\subsubsection{3D Cash-Or-Nothing Experiments}

\paragraph{Results}

In three dimensions (\(N=3\)), we retain the same nested meshes as above (\(21\) nodes per spatial direction on the coarse grid and \(41\) on the refined grid, with coincident-node inputs). The full parameter grid would be prohibitively large if a uniform discretization were used. Instead, we:
\begin{itemize}
    \item Choose reasonable continuous intervals for each parameter (e.g., \(\sigma_1,\sigma_2,\sigma_3\in[0.10,\,0.50]\), \(r\in[0,\,0.05]\), \(\rho_{12},\rho_{13},\rho_{23}\in[-0.99,\,0.99]\)), ensuring coverage of practically relevant regimes.
    \item Generate \(20\) random parameter tuples uniformly from these intervals (each tuple defines \((\sigma_1,\sigma_2,\sigma_3,r,\rho_{12},\rho_{13},\rho_{23})\)).
    \item Use \(m\in\{2,\,4,\,6,\,8,\,10,\,12,\,14,\,16,\,18\}\) of these \(20\) examples for training, and the remaining \(20 - m\) for testing.
    \item Neural network: 2 hidden layers of 20 neurons each (ReLU), Adam optimizer (\(\alpha=10^{-3}\)), 2000 epochs.
\end{itemize}

For each \(m\), we report the same refined-mesh and NNLCI RMSEs on the training and testing parameter sets.

\begin{table}[h]
    \centering
    \caption{3D Cash-Or-Nothing: Training RMSEs}
    \label{tab:3d_train}
    \begin{tabular}{ccc}
        \toprule
        \textbf{\# Training Examples} & \textbf{Refined RMSE (Train)} & \textbf{NNLCI RMSE (Train)} \\
        \midrule
        2  & $1.110603 \times 10^{1}$ & $2.910601 \times 10^{0}$ \\
        4  & $1.042734 \times 10^{1}$ & $2.537622 \times 10^{0}$ \\
        6  & $1.008995 \times 10^{1}$ & $2.430484 \times 10^{0}$ \\
        8  & $1.024061 \times 10^{1}$ & $2.915610 \times 10^{0}$ \\
        10 & $1.039161 \times 10^{1}$ & $2.330441 \times 10^{0}$ \\
        12 & $1.038717 \times 10^{1}$ & $2.263164 \times 10^{0}$ \\
        14 & $1.057045 \times 10^{1}$ & $2.364175 \times 10^{0}$ \\
        16 & $1.068657 \times 10^{1}$ & $2.876244 \times 10^{0}$ \\
        18 & $1.071019 \times 10^{1}$ & $2.541825 \times 10^{0}$ \\
        \bottomrule
    \end{tabular}
\end{table}

\begin{table}[h]
    \centering
    \caption{3D Cash-Or-Nothing: Testing RMSEs}
    \label{tab:3d_test}
    \begin{tabular}{ccc}
        \toprule
        \textbf{\# Training Examples} & \textbf{Refined RMSE (Test)} & \textbf{NNLCI RMSE (Test)} \\
        \midrule
        2  & $1.069643 \times 10^{1}$ & $2.322681 \times 10^{0}$ \\
        4  & $1.086713 \times 10^{1}$ & $2.435790 \times 10^{0}$ \\
        6  & $1.110237 \times 10^{1}$ & $2.924178 \times 10^{0}$ \\
        8  & $1.116501 \times 10^{1}$ & $3.414783 \times 10^{0}$ \\
        10 & $1.119138 \times 10^{1}$ & $2.623442 \times 10^{0}$ \\
        12 & $1.141710 \times 10^{1}$ & $2.863554 \times 10^{0}$ \\
        14 & $1.130530 \times 10^{1}$ & $2.560474 \times 10^{0}$ \\
        16 & $1.115970 \times 10^{1}$ & $2.907183 \times 10^{0}$ \\
        18 & $1.164521 \times 10^{1}$ & $2.612379 \times 10^{0}$ \\
        \bottomrule
    \end{tabular}
\end{table}

\bigskip
\noindent
\emph{Discussion:} Across all choices of \(m\), NNLCI reduces test RMSE from about \(10\)--\(11\) for the refined-mesh solver to about \(2.3\)--\(3.4\). Using as few as \(2\) training examples already yields a strong improvement; because the held-out test set then contains \(18\) random tuples, individual \(m\)-comparisons should be read with that variability in mind. Overall, the results indicate that NNLCI remains data-efficient in three spatial dimensions.

\subsection{Down-and-Out Barrier Call Options under the Heston Model}

\subsubsection{Model Formulation}

Under the Heston stochastic-volatility model, the PDE for a down-and-out barrier call option on a single asset with price \(S\) and instantaneous variance \(v\) is:
\[
\frac{\partial u}{\partial t} 
+ \tfrac{1}{2}v\,S^2\,\frac{\partial^2 u}{\partial S^2}
+ \rho\,\sigma\,v\,S\,\frac{\partial^2 u}{\partial S\,\partial v}
+ \tfrac{1}{2}\sigma^2\,v\,\frac{\partial^2 u}{\partial v^2}
+ r_d\,S\,\frac{\partial u}{\partial S}
+ \kappa(\eta - v)\,\frac{\partial u}{\partial v}
- r_d\,u = 0,
\]
for \((t,S,v)\in [0,T)\times [B,S_{\max}]\times [0,v_{\max}]\). The terminal payoff is
\[
u(T,S,v) = 
\begin{cases}
\max\{S - K,\,0\}, & \text{if }S(t) > B \;\forall\,t\in[0,T], \\
0, & \text{if }S(t)\le B \text{ for some } t<T,
\end{cases}
\]
where:
\begin{itemize}
    \item \(K=100\) is the strike price.
    \item \(B\in[80,95]\) is the barrier level: if \(S\) ever drops to \(B\) before maturity, the option is knocked out (pays zero).
    \item \(\kappa\in[0.5,\,5]\) is the mean-reversion rate of variance.
    \item \(\eta\in[0.01,\,0.2]\) is the long-run variance.
    \item \(\sigma\in[0.1,\,1]\) is the volatility of variance.
    \item \(\rho\in[-0.9,\,0.9]\) is the correlation between the Brownian motions for \(S\) and \(v\).
    \item \(r_d\in[0,\,0.10]\) is the domestic risk-free interest rate.
    \item We take maturity \(T=1\), spatial truncation \(S_{\max}=8K=800\), and \(v_{\max}=5\).
\end{itemize}

\subsubsection{Numerical Setup}

We discretize the Heston PDE with a Modified Craig–Sneyd ADI scheme on nested non-uniform rectangular grids in \((S,v)\), following the grid construction of in 't Hout and Foulon~\cite{hout2011adifinitedifferenceschemes}. The non-uniform meshes concentrate points near the strike and near \(v=0\), using clustering parameters \(c=K/5\) in the \(S\)-direction and \(d=v_{\max}/500\) in the \(v\)-direction. Denoting by \(m_1\) and \(m_2\) the numbers of mesh intervals in \(S\) and \(v\) (hence \(m_1+1\) and \(m_2+1\) nodes), we use
\begin{itemize}
    \item coarse mesh: \((m_1,m_2)=(50,25)\);
    \item refined mesh: \((m_1,m_2)=(100,50)\) (double resolution in each direction, so every coarse node coincides with a refined node);
    \item reference mesh: \((m_1,m_2)=(200,100)\), used only to generate training/testing labels.
\end{itemize}
Time is discretized with \(N=20\) steps of size \(\Delta t=T/N\). On \(S=B\) we impose the knock-out condition \(u=0\); the remaining boundary treatments follow the barrier-adapted conditions in~\cite{hout2011adifinitedifferenceschemes}.

Because the spatial mesh is non-uniform, each NNLCI input appends a local mesh-size metric
\[
h_{\text{local}} \;=\; \sqrt{\bigl(h_{\text{left}} + h_{\text{right}}\bigr)\,\bigl(h_{\text{up}} + h_{\text{down}}\bigr)}\,,
\]
where \(h_{\text{left}},\,h_{\text{right}}\) are the distances from the collocation point to its immediate neighbors in the \(S\)-direction, and \(h_{\text{up}},\,h_{\text{down}}\) are the corresponding distances in the \(v\)-direction. Collocation points lie on the coarse mesh; at each such point \((t_i,S_i,v_i)\) the network input is
\[
\bigl\{\,u^C(t_i,S_i,v_i),\;u^F(t_i,S_i,v_i),\;h_{\text{local}}(S_i,v_i)\bigr\},
\]
where \(u^F(t_i,S_i,v_i)\) is taken at the coincident refined-mesh node.

Parameter values used in the experiments:
\begin{itemize}
    \item \textbf{Barrier levels:} \(B\in\{80,\,85,\,90,\,95\}\).
    \item \textbf{Mean-reversion:} \(\kappa\in\{0.5,\,1.75,\,3.0,\,4.25,\,5.0\}\).
    \item \textbf{Long-run variance:} \(\eta\in\{0.01,\,0.06,\,0.11,\,0.16,\,0.20\}\).
    \item \textbf{Volatility of variance:} \(\sigma\in\{0.10,\,0.325,\,0.55,\,0.775,\,1.00\}\).
    \item \textbf{Correlation:} \(\rho\in\{-0.9,\,-0.45,\,0.0,\,0.45,\,0.9\}\).
    \item \textbf{Domestic rate:} \(r_d\in\{0,\,0.025,\,0.05,\,0.075,\,0.10\}\).
\end{itemize}
The five parameters \(\kappa,\eta,\sigma,\rho,r_d\) are each sampled at \(5\) uniformly spaced nodes (endpoints plus three interior points), while \(B\) takes the \(4\) listed values. The full Cartesian product therefore contains
\[
4\times 5^{5} \;=\; 12{,}500
\]
parameter tuples. Using the same training-gap convention as above---subsample every \(g\)-th node on each ordered parameter axis, then take the Cartesian product as the training set---we consider \(g\in\{2,\,4\}\). This gives
\[
g=2:\quad 2\times 3^{5} \;=\; 486
\quad\text{training tuples},\qquad
g=4:\quad 1\times 2^{5} \;=\; 32
\quad\text{training tuples,}
\]
with all remaining tuples used for testing. The network has two hidden layers of \(32\) neurons (ReLU), input dimension \(3\), and is trained with Adam (\(\alpha=10^{-3}\)) for \(2000\) epochs.

\subsubsection{Results}

Because no closed form is available, errors are measured against the reference-mesh solution. As in the cash-or-nothing tests, \(\mathrm{RMSE}_{\text{refined}}\) and \(\mathrm{RMSE}_{\text{NNLCI}}\) are aggregated over collocation points and over the training or testing parameter tuples.

\begin{table}[h]
    \centering
    \caption{Heston Model (Barrier Call): Training RMSEs}
    \label{tab:heston_train}
    \begin{tabular}{ccc}
        \toprule
        \textbf{Training Gap} & \textbf{Refined RMSE (Train)} & \textbf{NNLCI RMSE (Train)} \\
        \midrule
        2 & $1.141193 \times 10^{-2}$ & $5.083415 \times 10^{-4}$ \\
        4 & $1.861085 \times 10^{-2}$ & $3.867079 \times 10^{-4}$ \\
        \bottomrule
    \end{tabular}
\end{table}

\begin{table}[h]
    \centering
    \caption{Heston Model (Barrier Call): Testing RMSEs}
    \label{tab:heston_test}
    \begin{tabular}{ccc}
        \toprule
        \textbf{Training Gap} & \textbf{Refined RMSE (Test)} & \textbf{NNLCI RMSE (Test)} \\
        \midrule
        2 & $7.434591 \times 10^{-3}$ & $6.283897 \times 10^{-4}$ \\
        4 & $7.588650 \times 10^{-3}$ & $5.513930 \times 10^{-4}$ \\
        \bottomrule
    \end{tabular}
\end{table}

\bigskip
\noindent
\emph{Discussion:} On the test sets, NNLCI reduces RMSE by about \(12\times\) when \(g=2\) (\(7.43\times 10^{-3}\) to \(6.28\times 10^{-4}\)) and by about \(14\times\) when \(g=4\). Even with the sparser training set, the method substantially improves upon the refined-mesh solver.

\section{Conclusion and Future Work}
\label{sec:conclusion}

We have demonstrated that NNLCI can substantially reduce the numerical error of coarse-grid solutions in PDE-based option-pricing models with minimal training data. In both cash-or-nothing and barrier-call scenarios, NNLCI achieves a \(4\times\)–\(12\times\) RMSE reduction on test sets, even when the network is trained on only a small fraction of parameter combinations. In high-dimensional PDEs (e.g., 3-asset cash-or-nothing), training on as few as \(2\) examples out of \(20\) yielded the best performance, illustrating the method’s data efficiency.

Possible research directions include:
\begin{itemize}
    \item Testing larger local patches as NNLCI inputs---for example, 3 collocated points in 1D, $3\times 3$ patches in 2D, and $3\times 3\times 3$ patches in 3D---rather than the single-point (per mesh) inputs used in the present experiments. Because single-point inputs already perform well for this time‑irreversible equation, enlarging the local patch may yield more  substantial accuracy gains; multi‑point local patches of this type have been employed for time‑evolution equations such as the time‑reversible Maxwell’s equations~\cite{Cobb2023Maxwell}.
    \item Extending NNLCI to \emph{American-style} options (early exercise features), which require free-boundary solvers.
    \item Investigating \emph{multi-asset barrier options} under correlated stochastic-volatility models.
    \item Exploring \emph{reinforcement-learning} schemes that adaptively sample parameter space for maximal accuracy gain.
    \item Analyzing the theoretical properties of NNLCI in parabolic PDEs, particularly error bounds as a function of local patch size.
\end{itemize}

\bigskip

\bibliographystyle{unsrt}
\bibliography{reference}

\end{document}